\documentclass{article}
\usepackage[main,preprint]{neurips_2026}

\usepackage[utf8]{inputenc}
\usepackage[T1]{fontenc}
\usepackage{microtype}

\usepackage{hyperref}
\usepackage{url}
\usepackage{graphicx}
\usepackage{subcaption}
\usepackage{wrapfig}
\usepackage{xcolor}
\usepackage{booktabs}
\usepackage{tabularx}
\usepackage{multirow}
\usepackage{tikz}
\usetikzlibrary{arrows.meta,positioning,calc}

\usepackage{algorithm}
\usepackage[noend]{algorithmic}

\usepackage{enumitem}
\usepackage{amsmath}
\usepackage{amsfonts}
\usepackage{amssymb}
\usepackage{mathtools}
\usepackage{amsthm}
\usepackage{dsfont}
\mathtoolsset{showonlyrefs}
\setlist[enumerate]{label=(\roman*),leftmargin=25pt,topsep=0pt,partopsep=0pt,itemsep=0pt}
\setlist[itemize]{leftmargin=15pt,topsep=-2pt,partopsep=0pt,itemsep=2pt}

\theoremstyle{plain}
\newtheorem{theorem}{Theorem}[section]

\newtheorem{lemma}[theorem]{Lemma}

\theoremstyle{definition}

\newtheorem{assumption}[theorem]{Assumption}
\theoremstyle{remark}
\newtheorem{remark}[theorem]{Remark}

\DeclareMathOperator{\E}{\mathbb{E}}
\DeclareMathOperator{\R}{\mathbb{R}}
\DeclareMathOperator{\N}{\mathbb{N}}
\DeclareMathOperator{\KL}{KL}
\DeclareMathOperator{\loss}{\mathcal{L}}

\DeclareMathOperator{\probs}{\mathcal{P}}
\DeclareMathOperator{\obj}{\mathcal{S}}
\DeclareMathOperator*{\esssupraw}{ess\,sup}
\newcommand{\esssup}{\mathop{\smash[b]{\esssupraw}}\displaylimits}

\newcommand{\gh}{\href{https://github.com/mathefuchs/qdsb}{\texttt{github.com/mathefuchs/qdsb}}}

\title{QDSB: Quantized Diffusion Schrödinger Bridges}
\author{
  Tobias Fuchs \\
  Karlsruhe Institute of Technology, \\
  Karlsruhe, Germany \\
  \texttt{neurips@tobiasfuchs.de} \\
  \And
  Florian Kalinke \\
  Karlsruhe Institute of Technology, \\
  Karlsruhe, Germany \\
  \texttt{florian.kalinke@kit.edu} \\
  \And
  Nadja Klein \\
  Karlsruhe Institute of Technology, \\
  Karlsruhe, Germany \\
  \texttt{nadja.klein@kit.edu} \\
}

\begin{document}

\maketitle

\begin{abstract}
  Learning generative models in settings where the source and target distributions are only specified through unpaired samples is gaining in importance.
  Here, one frequently-used model are Schrödinger bridges (SB), which represent the most likely evolution between both endpoint distributions.
  To accelerate training, simulation-free SBs avoid the path simulation of the original SB models.
  However, learning simulation-free SBs requires paired data; a coupling of the source and target samples is obtained as the solution of the entropic optimal transport (OT) problem.
  As obtaining the optimal global coupling is infeasible in many practical cases,
  the entropic OT problem is iteratively solved on minibatches instead.
  Still, the repeated cost remains substantial and the locality can distort the global transport geometry.
  We propose \emph{quantized diffusion Schrödinger bridges} (QDSB), which compute the endpoint coupling on anchor-quantized endpoint distributions and lift the resulting plan back to original data points through cell-wise sampling.
  We show that the regularized optimal coupling is stable w.r.t.\ anchor quantization, with an error controlled by the quality of the anchor approximation.
  In real-world experiments, QDSB matches the sample quality of existing baselines, requiring substantially less time.
  Code and data are available at \gh.
\end{abstract}

\section{Introduction}
Generative modeling is increasingly used to learn the evolution of one data distribution to another,
and has found applications in image \citep{RombachBLEO22,ZhangHLG025}, audio \citep{KongPHZC21,LemercierRWMVG24}, and video synthesis \citep{BlattmannRLD0FK23,WangCMZHWYHYYGWSJCLDLQL25}.
Well-known approaches, such as denoising diffusion probabilistic models \citep{HoJA20}, score-based stochastic differential equations \citep{SongSKKEP21}, flow matching \citep{LipmanCBNL23}, rectified flows \citep{LiuGL23}, and consistency models \citep{SongDCS23}, train neural networks that represent a map evolving a simple \emph{reference} distribution to an observed data distribution.
In this setting, the reference distribution is typically available in closed form and the learned map allows generating new data.

Parallel to this setting, a core problem in generative modeling is learning a mapping from one observed \emph{data} distribution to another data distribution, naturally arising in unpaired image translation \citep{ZhuPIE17,LiuBK17} and in modeling cell-state dynamics \citep{0001HWDK20}.
Here, one observes samples from two endpoint distributions that are either paired or unpaired; we focus on the latter.
More precisely, we study the setting where one observes independent samples from a source and target distribution (both unknown), and the goal is to find the most likely evolution between both distributions, given a reference process.
Schrödinger bridges \citep{BortoliTHD21}, which connect diffusion-based models \citep{SongSKKEP21} with optimal transport (OT; \citealt{villani2009optimal,villani2021topics}), provide a principled framework to tackle this challenging problem.

However, estimating the Schrödinger bridge requires paired samples, that is, knowledge of which source sample should be mapped to which target sample.
In the unpaired case, this correspondence must be constructed from the data observed; the estimated pairing---or coupling---has key implications on the quality of the learned map.
While the original approach to learning the Schrödinger bridge involves simulating the data of intermediate evolution steps, which is computationally costly,
recent simulation-free methods make Schrödinger bridge training substantially more practical.
In particular, \citet{0001MFAZHWB24} showed that, by using the known decomposition of the Schrödinger bridge (with Brownian motion reference) into a weighted combination of Brownian bridges \citep{foellmer}, a so-called simulation-free estimation is possible;
their SF2M approach replaces costly simulation by score and flow matching against these Brownian bridges.

Yet, this simplification still requires paired samples.
To obtain a paired sample given an unpaired one, one solves the entropic OT problem \citep{Cuturi13} between the full source and target samples.
In practice, this is infeasible due to the size of the samples; the pairing is commonly estimated on minibatches only.
However, the minibatch heuristic may fail to capture the global transport geometry and can become overly diffuse under regularization \citep[Appendix~A.1]{0001MFAZHWB24}.
Further, the minibatch heuristic addresses the runtime problem only partially as the entropic OT problem is solved in every minibatch, rather than once globally.
As a result, the pairing step remains a computational and statistical bottleneck in simulation-free bridge training.

We address this bottleneck by proposing an anchor-based quantization for learning simulation-free Schrödinger bridges.
Our approach first quantizes each endpoint distribution onto a finite set of well-chosen anchors, then computes the entropic OT between the induced discrete measures, and finally samples original endpoints from matched anchor cells for bridge training.
Importantly, the simulation-free training objective itself remains unchanged, since the model is still supervised by Brownian bridges between raw data points rather than between anchor locations.
This yields a cheaper pairing procedure while allowing explicit control of the quality of the resulting OT approximation.

Our \textbf{contributions} are as follows.
\begin{itemize}
  \item \emph{Quantized couplings for simulation-free bridges.} We introduce an anchor-based coupling method for simulation-free Schrödinger bridge training from unpaired samples.
        The method computes entropic OT on quantized endpoint distributions and lifts the resulting plan back to original endpoint pairs through cell-wise sampling.
  \item \emph{Stability analysis.} We bound the influence of the perturbation of the solution of the entropic OT problem induced by the anchor quantization.
        The bound in Theorem~\ref{thm:main} and Lemma~\ref{thm:radius} provide a geometric principle that guides anchor selection in practice.
  \item \emph{Empirical evaluation.} We study the proposed coupling on toy and real-world tasks.
        The experiments indicate that our method preserves the quality of simulation-free bridge training while improving the time-performance trade-off of the coupling stage.
        All code and data is openly available at \gh, including code to reproduce all tables and figures.
\end{itemize}

\textbf{Outline.}
Section~\ref{sec:rel} discusses related work, while
Section~\ref{sec:methods} reviews simulation-free Schrödinger bridges and presents our quantized coupling method.
Section~\ref{sec:experiments} reports experiments, and Section~\ref{sec:conclusion} concludes.
All proofs are deferred to the appendices.

\section{Related work}
\label{sec:rel}
Instead of mapping one observed distribution to a simple reference distribution,
our work lies in the data-to-data setting, where the goal is to connect two observed endpoint distributions.
While non-SB-based methods exist in this setting \citep{ZhuPIE17,LiuBK17,0001HWDK20},
we focus our discussion on SB-based approaches.
Recent SB solvers include diffusion SB matching \citep{ShiBCD23}, generalized SB matching \citep{LiuLNKTC24}, and light and optimal SB matching \citep{GushchinKBK24}.
A closely related extension is LightSBB-M \citep{AlouadiHLMPT26}, which considers the Schrödinger-Bass bridge.
Among simulation-free approaches, the direct starting point of our work is SF2M \citep{0001MFAZHWB24}.
As detailed above, its remaining bottleneck is the computational time of solving an entropic OT problem in each training minibatch, which we address in this work.

Classical OT \citep{monge,kantorovich} provides the coupling foundation of both entropic OT and SBs, and entropic regularization makes OT practical through Sinkhorn iterations \citep{Cuturi13}.
However, large-scale OT remains expensive, which motivates approximations based on minibatches \citep{FatrasSFC21}, low-rank couplings \citep{ScetbonCP21}, or anchor points \citep{LinAD21}.
Recent work improves minibatch OT itself, for example, through unbalanced minibatch OT \citep{FatrasSFC21}, mini-batch partial OT \citep{NguyenNVPH22}, or hierarchical couplings between minibatches \citep{NguyenNNPBPLH22}.
Our method is orthogonal to these as they improve the minibatch OT surrogate, while leaving the surrounding training algorithm unchanged and thus still requiring the repeated solving of the entropic OT problem.
An alternative to reduce transport complexity is using a finite set of representatives \citep{LinAD21}, but then the inherent limitations of simulation-free training remain unchanged.
Our approach introduces an anchor-quantized coupling for the distinct problem of constructing endpoint couplings within simulation-free SB training rather than solving OT in general, and therefore has the benefit of not requiring the repeated solving of OT problems in each minibatch.

\section{Simulation-free Schrödinger bridges with quantized couplings}
\label{sec:methods}
This section introduces QDSB, our quantized coupling procedure for simulation-free SB training from unpaired samples.
To set the stage, we recall the simulation-free objective of \citet{0001MFAZHWB24}.

\subsection{Preliminaries}
\label{sec:prelim}
\textbf{Notation.}
Let $d \in \N$, equip $\R^d$ with a fixed norm $\|\cdot\|$, and let $c(x, y) = \|x - y\|$ denote the associated distance function.
Given $a \in [1, \infty)$, let $\probs_a(\R^d)$ denote the set of Borel probability measures $\mu$ on $\R^d$ with finite $a$-th moment,
that is, $\int c(x, y)^a \,\mathrm{d} \mu(x) < \infty$ for $y \in \R^d$.
For $a = \infty$, $\probs_\infty(\R^d)$ is the set of measures with bounded support.
For $\mu, \nu \in \probs_a(\R^d)$, let $\Pi(\mu, \nu)$ denote the set of couplings with marginals $\mu$ and $\nu$,
and $W_a(\mu, \nu)$ denote their associated $a$-Wasserstein distance:
for $a \in [1, \infty)$, $W_a(\mu, \nu)^a = \inf_{\pi \in \Pi(\mu, \nu)} \int c(x, y)^a \,\mathrm{d} \pi(x, y)$;
for $a = \infty$, $W_\infty(\mu, \nu) = \inf_{\pi \in \Pi(\mu, \nu)} \esssup_{(x, y) \sim \pi} c(x, y)$.
For a Borel measurable map $T : \R^d \to \R^d$ and probability measure $\mu$, we write $T_{\#}\mu$ for the pushforward of $\mu$ under $T$.

\textbf{Schrödinger bridge (SB).}
Fix $d \in \N$, $a \in [1, \infty]$, and let $q_0, q_1 \in \probs_a(\R^d)$.
The goal of the SB problem is to find the most likely evolution between $q_0$ and $q_1$, that is,
it is the solution $P^* = \arg\min_{P} \KL( P \| Q )$ subject to $p_0=q_0$ and $p_1=q_1$, where $P$ is a probability measure on continuous $[0, 1] \to \R^d$ functions that is absolutely continuous w.r.t.\ a given reference measure $Q$, and law $p$ with $p_t$ denoting the time‑$t$ marginal of~$P$ with $t \in [0, 1]$.

\textbf{Simulation-free training (SF2M).}
\citet{0001MFAZHWB24} consider the Brownian reference measure $Q$ given by the law of the scaled multivariate Brownian motion $\sigma B$, where $B$ is a standard Brownian motion and $\sigma > 0$.
In that case, $P^*$ admits a representation as a mixture of Brownian bridges
\begin{align}
  P^*((x_t)_{t \in [0, 1]}) = \int Q((x_t)_{t \in [0, 1]} \mid x_0, x_1) \,\mathrm{d} \pi^{2\sigma^2}_{q_0,q_1}(x_0, x_1) \text{,} \label{eq:brownian-mix}
\end{align}
where $Q(\cdot \mid x_0, x_1)$ is the Brownian bridge connecting $x_0$ and $x_1$.
It is weighted by the entropic OT coupling $\pi^{2\sigma^2}_{q_0,q_1}$ between $q_0$ and $q_1$, that is, $\pi^\tau_{q_0,q_1}$ is the minimizer of
\begin{align}
  \obj_\tau(q_0, q_1) = \inf_{\pi \in \Pi(q_0, q_1)} \left[ \int c(x_0, x_1) \,\mathrm{d} \pi(x_0, x_1) + \tau \KL(\pi \| q_0 \otimes q_1) \right] \text{,}
  \label{eq:entropic-ot}
\end{align}
where $\tau = 2\sigma^2$ in~\eqref{eq:brownian-mix}.
In this case, it is known \citep{0001MFAZHWB24} that to approximate $P^*$ it suffices to learn the so-called drift $u_t^\circ(x \mid x_0, x_1) = \frac{1 - 2t}{t(1-t)} (x - (tx_1 + (1 - t) x_0)) + (x_1 - x_0)$ and score $\nabla_x \log p_t(x \mid x_0, x_1) = \frac{tx_1 + (1-t) x_0 - x}{\sigma^2 t(1-t)}$ of the underlying stochastic differential equation.
Indeed, \citet{0001MFAZHWB24} then fit neural networks approximating each, that is, they minimize
\begin{align}
  \!\loss_\theta \!=\! \E_{t \sim \mathcal{U}(0, 1), z \sim q(z), x \sim p_t(x \mid z)} \!\!\left[ \| v_{\theta}(t, x) \!-\! u_t^\circ(x \mid z) \|^2 \!+\! \lambda(t)^2 \| s_{\theta}(t, x) \!-\! \nabla_x \log p_t(x \mid z) \|^2 \right] \!\text{,}\quad \label{eq:loss}
\end{align}
with $z = (x_0, x_1)$, $v_{\theta}(t, x)$ a neural network approximating the drift term, $s_{\theta}(t, x)$ a neural network approximating the score term, $\theta$ the network parameters, and $\lambda(t) > 0$ a weighting function.

Since only unpaired i.i.d.\ samples from $q_0$ and $q_1$ are observed, the endpoint coupling $q(x_0, x_1)$ is unknown and must be constructed.
\citet{0001MFAZHWB24} achieve this by solving a separate OT problem on each training minibatch and use the resulting matches as samples $z \sim q(x_0, x_1)$.
However, minibatch OT is only a surrogate for the desired global coupling: subsampled estimators need not preserve the exact marginals, and when combined with entropic regularization they can produce overly dense, nearly uniform transport plans that obscure the data geometry \citep[Appendix~A.1]{0001MFAZHWB24}.
The quality of the learned bridge critically depends on how $q(x_0, x_1)$ is constructed.
This observation motivates our improved coupling procedure proposed in the following.

\subsection{QDSB: Quantized diffusion Schrödinger bridges}

\begin{wrapfigure}[19]{r}{0pt}
  \centering
  \includegraphics[width=2.35in]{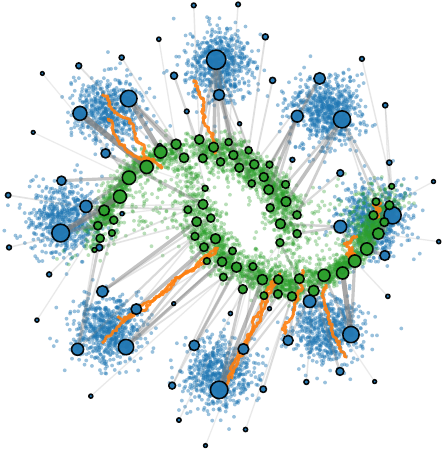}
  \caption{Illustration of QDSB on 2D toy datasets, see Section~\ref{sec:experiments}.}
  \label{fig:anchors}
\end{wrapfigure}

We replace the minibatch OT heuristic in \citet{0001MFAZHWB24} with a coupling computed on a quantized representation of the endpoint distributions.
For each endpoint distribution $q_i$, $i \in \{0, 1\}$, we select a finite anchor set $A_i \subseteq \R^d$ and define a map $T_i : \R^d \to A_i$ that assigns each point to its closest anchor.
This induces the discrete pushforward measure $\tilde{q}_i := (T_i)_{\#}q_i$.
We then solve the entropic OT problem once between $\tilde{q}_0$ and $\tilde{q}_1$ and use the resulting anchor-level OT plan to sample endpoint pairs for the loss in~\eqref{eq:loss}.

While we explain this procedure more thoroughly in Section~\ref{sec:algo},
Figure~\ref{fig:anchors} briefly illustrates this construction.
The large blue and green points denote the source and target anchors, and the gray links denote the OT plan between them.
During training, endpoint pairs are sampled from matched anchor cells.
The anchors and transport links are weighted to match the empirical marginals of the original endpoint distributions.
The orange curves indicate example trajectories.

The networks are still trained on original endpoint samples rather than on anchor coordinates.
At the empirical level, let $C_i(\alpha)$ denote the set of training samples assigned to anchor $\alpha \in A_i$, and let $\hat{q}_i(\cdot \mid \alpha)$ be the uniform distribution on that cell.
Each anchor $\alpha \in A_i$ therefore carries mass $|C_i(\alpha)| / n$, and the anchor-level OT problem is solved between weighted anchor (and thus discrete) measures rather than between uniform measures on $A_0$ and $A_1$.
The anchor-level transport plan $\pi^\tau_{\tilde{q}_0,\tilde{q}_1}$ induces the coupling
$q_{\mathrm{anc}}(x_0, x_1) = \sum_{\alpha \in A_0} \sum_{\beta \in A_1} \pi^\tau_{\tilde{q}_0, \tilde{q}_1}\!(\alpha, \beta)\, \hat{q}_0(x_0 \mid \alpha)\, \hat{q}_1(x_1 \mid \beta)$.
By construction, $q_{\mathrm{anc}}$ has the empirical marginals of the two endpoint datasets; the approximation therefore affects only how source and target points are paired, not which endpoint samples are seen.

This construction separates the pairing step from the sample generation step used during training.
The anchor‑level transport plan determines which source cells are matched to which target cells, while sampling inside these cells provides original endpoint pairs for the Brownian-bridge supervision.
As a result, the expensive OT computation is moved outside the inner optimization loop and solved on a much smaller discrete problem.
The next two subsections justify this.

\subsubsection{Theory: Anchor quantization stability}
The central approximation underlying QDSB is that the entropic optimal transport problem is solved after first quantizing the endpoint distributions onto finite anchor sets.
The key theoretical question is therefore whether the resulting anchor-level transport plan remains sufficiently close to the transport plan that would be obtained from the unquantized marginals.
In the following, we answer this question in the affirmative with only mild assumptions on the transport cost and the regularized transport problem.
Specifically, Assumption~\ref{thm:a1} controls the sensitivity of the transport cost to perturbations of the marginals (admissibility).
The admissibility requirement is mild and includes standard choices such as the squared Euclidean cost.
Assumption~\ref{thm:a2} is the transport inequality of \citet{EcksteinN22} that relates the stability of the objective values to the stability of optimal couplings.
Similarly, this requirement is also mild and satisfied in the common case of $a = 2$ and sub-Gaussian marginals (see discussion under \citep[Lemma~3.10]{EcksteinN22}).

\begin{assumption}[Admissible cost]
  \label{thm:a1}
  Let $a \in [1,\infty]$.
  The cost function $c : \R^d \times \R^d \to [0, \infty)$ is Borel measurable and satisfies the admissibility condition of \citet[Definition~3.3]{EcksteinN22} with constant $L \geq 0$, that is, for any probability measures $r_0, r'_0, r_1, r'_1 \in \probs_a(\R^d)$,
  and any couplings $\pi \in \Pi(r_0, r_1)$, $\pi' \in \Pi(r_0', r_1')$, it holds that
  \begin{align}
    \left| \int c \,\mathrm{d}( \pi - \pi' ) \right| \leq L W'_a(\pi, \pi') \text{ with }
    W'_a(\pi, \pi') := \begin{cases}
                         \left( W_a(r_0, r'_0)^a + W_a(r_1, r'_1)^a \right)^{\frac{1}{a}} & \text{if } a < \infty \text{,} \\
                         \max \{ W_a(r_0, r'_0), W_a(r_1, r'_1) \}                        & \text{if } a = \infty \text{.}
                       \end{cases}
    \notag
  \end{align}
\end{assumption}

\begin{assumption}[Transport inequality]
  \label{thm:a2}
  Let $a \in [1, \infty]$ and $b \in [1, \infty)$ with $b \leq a$.
  The endpoint distributions $q_0, q_1 \in \probs_a(\R^d)$ satisfy the transport inequality Eq.~($I_q$) of \citet{EcksteinN22} with constant $C_b \geq 0$, that is, $W'_{b}(\pi, \pi') \leq C_{b} \KL(\pi, \pi')^{1/(2b)}$ for all $\pi, \pi' \in \Pi(q_0, q_1)$.
\end{assumption}

Assumptions~\ref{thm:a1} and~\ref{thm:a2} provide the ingredients needed to control the effect of anchor quantization on the transport problem in~\eqref{eq:entropic-ot}.
The first assumption bounds the perturbation of the objective value induced by changes in the marginals, while the second converts this value control into a bound on the corresponding optimal couplings.
Under these assumptions, Theorem~\ref{thm:main} shows that the resulting approximation error scales directly with the quantization error of the anchor projections.

\begin{theorem}[Anchor stability]
  \label{thm:main}
  Let $i \in \{0, 1\}$, $a \in [1, \infty]$, $q_i \in \probs_a(\R^d)$, and $\tilde{q}_i := (T_i)_{\#}q_i$, where $T_i : \R^d \to A_i$ are measurable projections onto finite anchor sets $A_i \subseteq \R^d$.
  Define the quantization error as $\varepsilon_i := ( \int \left\| x_i - T_i(x_i) \right\|^a \,\mathrm{d}q_i(x_i) )^{1/a}$ if $a < \infty$,
  and $\varepsilon_i := \esssup_{x_i \sim q_i} \| x_i - T_i(x_i) \|$ if $a = \infty$.
  Set $\Delta_a := (\varepsilon_0^a + \varepsilon_1^a)^{1/a}$ if $a < \infty$, and $\Delta_a := \max \{ \varepsilon_0, \varepsilon_1 \}$ if $a = \infty$.
  Recall the entropic OT objective $\obj_\tau$ from~\eqref{eq:entropic-ot}.
  Let $\pi^\tau_{q_0,q_1}$ and $\pi^\tau_{\tilde{q}_0,\tilde{q}_1}$ denote the corresponding optimal couplings w.r.t.\ $\obj_\tau(q_0,q_1)$ and $\obj_\tau(\tilde{q}_0,\tilde{q}_1)$, respectively.
  Further, assume that $c : \R^d \times \R^d \to \R$ is Borel measurable and admissible with constant $L \geq 0$ (Assumption~\ref{thm:a1}), and that the transport inequality in Assumption~\ref{thm:a2} holds for $(q_0, q_1)$ with $C_b \geq 0$.
  Then, the following bounds hold:
  \begin{enumerate}
    \item \textbf{Endpoint perturbation bound:} $W_a(q_0, \tilde{q}_0) \leq \varepsilon_0$ and $W_a(q_1, \tilde{q}_1) \leq \varepsilon_1$.
    \item \textbf{Entropic value stability:} $| \obj_\tau(q_0, q_1) - \obj_\tau(\tilde{q}_0, \tilde{q}_1) | \leq L \Delta_a$. 
    \item \textbf{Entropic coupling stability:} $W'_{b}(\pi^\tau_{q_0,q_1}, \pi^\tau_{\tilde{q}_0,\tilde{q}_1}) \leq 2^{1/b - 1/a} \Delta_a + C_{b}(2 L \Delta_a / \tau)^{1/(2b)}$.
  \end{enumerate}
\end{theorem}

Theorem~\ref{thm:main} has three practical implications.
First, quantizing each endpoint distribution only moves mass by at most the quantization error $\varepsilon_i$.
Second, the entropic OT value changes at most linearly in the combined error scale $\Delta_a$.
Third, the Wasserstein distance between the optimal regularized coupling itself and its quantized version is bounded.
As a result, the anchor‑level transport plan remains close to the original transport plan whenever the anchor cells are sufficiently small.

Lemma~\ref{thm:radius} turns the abstract quantity $\Delta_a$ into a geometric design principle for choosing anchors.
If every point lies within radius $r_i$ of its assigned anchor, then all stability terms are controlled by these radii.
Consequently, anchor selection should emphasize coverage of the endpoint distributions rather than focusing exclusively on high-density regions.
This is exactly the metric k-center problem \citep{kmetric}:
Given a finite set of points $V \subseteq \R^d$ and number of anchors $k$, its goal is to find a set $C \subseteq V$ with $|C| = k$ that minimizes the coverage radius $r^* = \max_{u \in V} \min_{v \in C} c(u, v)$.
In other words, for all points $x \in V$ and their associated anchors $y_x \in C$, it holds that $c(x, y_x) \leq r^*$.
The greedy farthest-first traversal algorithm approximately solves the $k$-center problem and has an approximation factor of 2, that is, $r \leq 2 r^*$ (elaborated in Theorem~\ref{thm:kmetric}).
We note that a 2-approximation is the best possible approximation with polynomial runtime.

\begin{lemma}[Radius-based bound]
  \label{thm:radius}
  Assume the setting of Theorem~\ref{thm:main}.
  Suppose that, for each $i \in \{0, 1\}$, the anchor projection $T_i$ satisfies $\| x_i - T_i(x_i) \| \leq r_i$ for $q_i$-almost every $x_i$. Then, $(i)$ $\varepsilon_i \leq r_i$, and $(ii)$ $\Delta_a \leq (r_0^a + r_1^a)^{1/a}$ if $a < \infty$, and $\Delta_a \leq \max \{ r_0, r_1 \}$ if $a = \infty$.
\end{lemma}

If the projections $T_i$ satisfy the radius condition in Lemma~\ref{thm:radius}, the bounds in Theorem~\ref{thm:main} can be read directly in terms of geometric coverage of the endpoint distributions.

\begin{remark}
  \label{remark:radius}
  Lemma~\ref{thm:radius}~$(i)$ and~$(ii)$ can be substituted directly into Theorem~\ref{thm:main}~$(i)$\,--\,$(iii)$.
  Hence, once the anchor sets $A_0$ and $A_1$ induce $T_1$ and $T_2$ achieving the desired coverage radii, stability of the quantized couplings is solely controlled by the granularity of the anchor discretization.
\end{remark}

At the same time, the usual limitations of high-dimensional metric quantization arise:
if distances become less informative, obtaining small coverage radii may require many anchors.
Nevertheless, our empirical results, which include an image translation experiment in a 512-dimensional latent space, indicate that the approximation remains effective in practice.
One reason is that the anchor-level transport plan only determines which source and target cells interact, while the Brownian bridge supervision is still computed from original endpoints sampled within these cells.
Hence, when the cells are sufficiently small, re-pairing points within a cell does not substantially change the training.

\subsubsection{Algorithm}
\label{sec:algo}
Algorithm~\ref{alg:algo} summarizes the proposed training procedure.
It consists of an outer coupling stage, which constructs and periodically refreshes the anchor‑level transport plan, and an inner training stage, which reuses this plan to generate endpoint pairs for the simulation-free loss.

\begin{algorithm}[t]
  \caption{QDSB: Quantized diffusion Schrödinger bridges.}
  \label{alg:algo}
  \begin{algorithmic}[1]
    \item[\textbf{Input:}] Samples $\{x_0^{(i)}\}_{i=1}^n$ of source distribution $q_0$, samples $\{x_1^{(i)}\}_{i=1}^n$ of target distribution $q_1$, anchor set cardinality $k \in \mathbb{N}$, anchor refresh interval $t_{\mathrm{refresh}} \in \mathbb{N}$;
    \item[\textbf{Output:}] Trained score network $s_\theta$ and drift network $v_\theta$;
    \vspace{0.2em}
    \STATE \emph{$\triangleright$ Anchor construction}
    \STATE $A_0, A_1 \leftarrow $ Select $k$ points each from $\{x_0^{(i)}\}_{i=1}^n$ and $\{x_1^{(i)}\}_{i=1}^n$ using farthest-first traversal;
    \vspace{0.3em}
    \STATE Compute empirical OT plan $\pi^\tau_{\tilde{q}_0,\tilde{q}_1}$ between the anchor measures induced by $A_0$ and $A_1$;
    \vspace{0.4em}
    \STATE \emph{$\triangleright$ Training}
    \vspace{0.1em}
    \STATE Initialize score network $s_\theta$ and drift network $v_\theta$;
    \vspace{0.1em}
    \WHILE{training}
    \vspace{0.1em}
    \STATE Refresh anchor sets $A_0$, $A_1$, and anchor coupling $\pi^\tau_{\tilde{q}_0,\tilde{q}_1}$ every $t_{\mathrm{refresh}}$ epochs (see lines~1--3);
    \STATE Sample $t \in [0, 1]$ and an anchor pair $(\tilde{\alpha}, \tilde{\beta})\sim \pi^\tau_{\tilde{q}_0,\tilde{q}_1}$;
    \STATE Sample \(\tilde x_0 \sim \hat q_0(\cdot \mid \tilde \alpha)\),
    \(\tilde x_1 \sim \hat q_1(\cdot \mid \tilde \beta)\), set \(z \leftarrow (\tilde x_0,\tilde x_1)\), and sample $x \sim p_t(x \mid z)$; 
    \vspace{0.3em}
    \STATE Compute \eqref{eq:loss}, that is, $\loss_\theta \leftarrow \| v_{\theta}(t, x) - u_t^\circ(x \mid z) \|^2 + \lambda(t)^2 \| s_{\theta}(t, x) - \nabla \log p_t(x \mid z) \|^2$;
    \vspace{0.3em}
    \STATE Update $\theta$ using backpropagation on $\loss_\theta$;
    \ENDWHILE
  \end{algorithmic}
\end{algorithm}

The \textbf{first stage} constructs the anchor representation of the two endpoint datasets.
We select anchors independently on source and target data via farthest-first traversal, which is motivated in Lemma~\ref{thm:radius} and detailed in Algorithm~\ref{alg:farthest-first} in the appendix.
Given the anchor sets, nearest-anchor assignment induces source cells $C_0(\alpha)$ for $\alpha \in A_0$ and target cells $C_1(\beta)$ for $\beta \in A_1$, together with the corresponding empirical measures supported on the anchors and used for solving the (entropic) OT problem.

The \textbf{second stage} amortizes the transport plan computation across many gradient updates.
Instead of solving an OT problem for each training minibatch, QDSB computes a discrete entropic OT plan on the anchor supports only and refreshes it every $t_{\mathrm{refresh}}$ epochs.
This substantially reduces the computational overhead while still allowing the anchor geometry and coupling to adapt over the course of training.
Within the training loop, the anchor-level transport plan is lifted back to pairs of original data points.
Sampling $(\tilde{\alpha}, \tilde{\beta})$ first chooses a matched pair of source and target cells, and sampling uniformly from $C_0(\tilde{\alpha})$ and $C_1(\tilde{\beta})$ then restores the empirical variability that would be lost if we trained on anchors alone.
This step is crucial in practice, since the score and drift supervision are computed from Brownian bridges between actual endpoints rather than between anchors.
Given $z = (\tilde{x}_0, \tilde{x}_1)$, an intermediate state is sampled from the associated Brownian bridge, and the networks are trained to regress toward the closed-form drift and score.
Increasing the number of anchors refines the geometry captured by the transport plan and recovers the original objective in the limit.

\subsection{Positioning w.r.t.\ minibatch OT}
\citet{0001MFAZHWB24} use minibatch OT to approximate the endpoint coupling, making it natural to contrast this approach with QDSB.
Minibatch OT replaces $q_0$ and $q_1$ by uniform empirical measures on random subsamples and solves OT only on these supports.
\citet{FatrasSFC21} analyze the resulting averaged surrogate and Monte Carlo estimator.
The connection to QDSB becomes explicit when minibatches are interpreted as random anchor sets $A_0$ and $A_1$.
Given these sampled supports, Theorem~\ref{thm:main} and Lemma~\ref{thm:radius} show that the induced coupling error is controlled by the corresponding quantization scale $\Delta_a$, and equivalently by the covering radii of $A_0$ and $A_1$.
The key differences are that QDSB selects anchors to explicitly minimize this radius and constructs weighted pushforward measures $(T_i)_{\#} q_i$ rather than uniform minibatch weights.
The resulting stability guarantees make QDSB more suitable than random minibatch OT approximations for simulation‑free SB training.

\begin{table}[t]
  \centering
  \caption{MMD results on three toy settings (Table~\ref{tab:results-2d}) and three real-world datasets (Table~\ref{tab:results-rw}). We report mean and standard deviation over five seeds, and highlight the best value in each column.}
  \label{tab:results}
  \begin{subtable}{\textwidth}
    \centering
    \caption{MMD results on 2D toy experiments. Lower values are better.}
    \label{tab:results-2d}
    {\setlength{\tabcolsep}{3.2pt}
      \begin{small}
        \begin{tabularx}{\textwidth}{@{}>{\raggedright\arraybackslash}Xccc|ccc|cccc@{}}
          \toprule
          \multirow[c]{2}{*}[-0.8mm]{Method} & \multicolumn{3}{c}{After 10\,s}             & \multicolumn{3}{c}{After 60\,s}             & \multicolumn{4}{c}{500 epochs + total time}                                                                                                                                                                                                                                                                                                                                  \\
          \cmidrule(lr){2-4}\cmidrule(lr){5-7}\cmidrule(ll){8-11}
                                             & 8G$\to$M                                    & G$\to$M                                     & G$\to$8G                                    & 8G$\to$M                                    & G$\to$M                                     & G$\to$8G                                    & 8G$\to$M                                    & G$\to$M                                     & G$\to$8G                                    & Time                                       \\
          \midrule
          \shortstack[l]{QDSB\\(ours)}       & \shortstack[c]{\textbf{0.024}\\$\pm$ 0.007} & \shortstack[c]{\textbf{0.015}\\$\pm$ 0.010} & \shortstack[c]{\textbf{0.015}\\$\pm$ 0.015} & \shortstack[c]{\textbf{0.019}\\$\pm$ 0.007} & \shortstack[c]{\textbf{0.004}\\$\pm$ 0.005} & \shortstack[c]{\textbf{0.010}\\$\pm$ 0.016} & \shortstack[c]{0.023\\$\pm$ 0.010}          & \shortstack[c]{0.010\\$\pm$ 0.009}          & \shortstack[c]{\textbf{0.010}\\$\pm$ 0.016} & \multirow[c]{1}{*}[0.5em]{105\,s}          \\[0.3em]
          \shortstack[l]{DSB\\(2021)}        & \shortstack[c]{0.181\\$\pm$ 0.022}          & \shortstack[c]{0.240\\$\pm$ 0.041}          & \shortstack[c]{0.208\\$\pm$ 0.010}          & \shortstack[c]{0.238\\$\pm$ 0.015}          & \shortstack[c]{0.165\\$\pm$ 0.018}          & \shortstack[c]{0.261\\$\pm$ 0.013}          & \shortstack[c]{0.202\\$\pm$ 0.014}          & \shortstack[c]{0.168\\$\pm$ 0.023}          & \shortstack[c]{0.259\\$\pm$ 0.033}          & \multirow[c]{1}{*}[0.5em]{\textbf{103\,s}} \\[0.3em]
          \shortstack[l]{DSBM\\(2023)}       & \shortstack[c]{0.081\\$\pm$ 0.005}          & \shortstack[c]{0.049\\$\pm$ 0.005}          & \shortstack[c]{0.103\\$\pm$ 0.025}          & \shortstack[c]{0.182\\$\pm$ 0.046}          & \shortstack[c]{0.083\\$\pm$ 0.025}          & \shortstack[c]{0.061\\$\pm$ 0.021}          & \shortstack[c]{0.139\\$\pm$ 0.029}          & \shortstack[c]{0.128\\$\pm$ 0.027}          & \shortstack[c]{0.103\\$\pm$ 0.039}          & \multirow[c]{1}{*}[0.5em]{193\,s}          \\[0.3em]
          \shortstack[l]{SF2M\\(2024)}       & \shortstack[c]{0.040\\$\pm$ 0.007}          & \shortstack[c]{0.028\\$\pm$ 0.004}          & \shortstack[c]{0.041\\$\pm$ 0.009}          & \shortstack[c]{0.024\\$\pm$ 0.008}          & \shortstack[c]{0.020\\$\pm$ 0.005}          & \shortstack[c]{\textbf{0.010}\\$\pm$ 0.017} & \shortstack[c]{\textbf{0.022}\\$\pm$ 0.016} & \shortstack[c]{\textbf{0.006}\\$\pm$ 0.008} & \shortstack[c]{0.011\\$\pm$ 0.016}          & \multirow[c]{1}{*}[0.5em]{205\,s}          \\[0.3em]
          \shortstack[l]{SF2M\\+ mPOT}       & \shortstack[c]{0.109\\$\pm$ 0.008}          & \shortstack[c]{0.202\\$\pm$ 0.008}          & \shortstack[c]{0.108\\$\pm$ 0.004}          & \shortstack[c]{0.092\\$\pm$ 0.008}          & \shortstack[c]{0.188\\$\pm$ 0.009}          & \shortstack[c]{0.063\\$\pm$ 0.006}          & \shortstack[c]{0.090\\$\pm$ 0.012}          & \shortstack[c]{0.181\\$\pm$ 0.007}          & \shortstack[c]{0.057\\$\pm$ 0.006}          & \multirow[c]{1}{*}[0.5em]{258\,s}          \\[0.3em]
          \shortstack[l]{LightSB-M\\(2024)}  & \shortstack[c]{0.669\\$\pm$ 0.000}          & \shortstack[c]{0.633\\$\pm$ 0.002}          & \shortstack[c]{0.673\\$\pm$ 0.001}          & \shortstack[c]{0.671\\$\pm$ 0.001}          & \shortstack[c]{0.620\\$\pm$ 0.002}          & \shortstack[c]{0.669\\$\pm$ 0.001}          & \shortstack[c]{0.644\\$\pm$ 0.009}          & \shortstack[c]{0.225\\$\pm$ 0.006}          & \shortstack[c]{0.603\\$\pm$ 0.004}          & \multirow[c]{1}{*}[0.5em]{550\,s}          \\
          \bottomrule
        \end{tabularx}
      \end{small}
    }
  \end{subtable}
  \vspace{0.4em}
  \begin{subtable}{\textwidth}
    \centering
    \caption{MMD results on real-world cellular biology experiments. Lower values are better.}
    \label{tab:results-rw}
    {\setlength{\tabcolsep}{3.2pt}
      \begin{small}
        \begin{tabularx}{\textwidth}{@{}>{\raggedright\arraybackslash}Xccc|ccc|cccc@{}}
          \toprule
          \multirow[c]{2}{*}[-0.8mm]{Method} & \multicolumn{3}{c}{After 10\,s}             & \multicolumn{3}{c}{After 60\,s}             & \multicolumn{4}{c}{1000 epochs + total time}                                                                                                                                                                                                                                                                                                                                 \\
          \cmidrule(lr){2-4}\cmidrule(lr){5-7}\cmidrule(ll){8-11}
                                             & EB                                          & Cite                                        & Multi                                        & EB                                          & Cite                                        & Multi                                       & EB                                          & Cite                                        & Multi                                       & Time                                      \\
          \midrule
          \shortstack[l]{QDSB\\(ours)}       & \shortstack[c]{\textbf{0.198}\\$\pm$ 0.023} & \shortstack[c]{\textbf{0.183}\\$\pm$ 0.004} & \shortstack[c]{\textbf{0.224}\\$\pm$ 0.054}  & \shortstack[c]{\textbf{0.193}\\$\pm$ 0.026} & \shortstack[c]{\textbf{0.179}\\$\pm$ 0.003} & \shortstack[c]{0.231\\$\pm$ 0.060}          & \shortstack[c]{0.193\\$\pm$ 0.026}          & \shortstack[c]{0.183\\$\pm$ 0.005}          & \shortstack[c]{0.232\\$\pm$ 0.061}          & \multirow[c]{1}{*}[0.5em]{\textbf{74\,s}} \\[0.3em]
          \shortstack[l]{DSB\\(2021)}        & \shortstack[c]{0.264\\$\pm$ 0.093}          & \shortstack[c]{0.195\\$\pm$ 0.034}          & \shortstack[c]{0.295\\$\pm$ 0.084}           & \shortstack[c]{0.254\\$\pm$ 0.053}          & \shortstack[c]{0.219\\$\pm$ 0.039}          & \shortstack[c]{0.313\\$\pm$ 0.146}          & \shortstack[c]{0.221\\$\pm$ 0.029}          & \shortstack[c]{0.245\\$\pm$ 0.044}          & \shortstack[c]{0.318\\$\pm$ 0.140}          & \multirow[c]{1}{*}[0.5em]{103\,s}         \\[0.3em]
          \shortstack[l]{DSBM\\(2023)}       & \shortstack[c]{0.233\\$\pm$ 0.031}          & \shortstack[c]{0.198\\$\pm$ 0.056}          & \shortstack[c]{0.247\\$\pm$ 0.033}           & \shortstack[c]{0.221\\$\pm$ 0.055}          & \shortstack[c]{0.189\\$\pm$ 0.029}          & \shortstack[c]{0.230\\$\pm$ 0.018}          & \shortstack[c]{0.219\\$\pm$ 0.054}          & \shortstack[c]{0.189\\$\pm$ 0.025}          & \shortstack[c]{0.231\\$\pm$ 0.024}          & \multirow[c]{1}{*}[0.5em]{92\,s}          \\[0.3em]
          \shortstack[l]{SF2M\\(2024)}       & \shortstack[c]{0.214\\$\pm$ 0.021}          & \shortstack[c]{0.185\\$\pm$ 0.016}          & \shortstack[c]{0.229\\$\pm$ 0.051}           & \shortstack[c]{0.197\\$\pm$ 0.019}          & \shortstack[c]{\textbf{0.179}\\$\pm$ 0.008} & \shortstack[c]{0.223\\$\pm$ 0.052}          & \shortstack[c]{0.190\\$\pm$ 0.028}          & \shortstack[c]{0.180\\$\pm$ 0.007}          & \shortstack[c]{0.232\\$\pm$ 0.062}          & \multirow[c]{1}{*}[0.5em]{318\,s}         \\[0.3em]
          \shortstack[l]{SF2M\\+ mPOT}       & \shortstack[c]{0.216\\$\pm$ 0.032}          & \shortstack[c]{0.186\\$\pm$ 0.014}          & \shortstack[c]{\textbf{0.224}\\$\pm$ 0.031}  & \shortstack[c]{0.215\\$\pm$ 0.031}          & \shortstack[c]{0.186\\$\pm$ 0.007}          & \shortstack[c]{\textbf{0.222}\\$\pm$ 0.036} & \shortstack[c]{0.206\\$\pm$ 0.039}          & \shortstack[c]{0.185\\$\pm$ 0.003}          & \shortstack[c]{\textbf{0.226}\\$\pm$ 0.045} & \multirow[c]{1}{*}[0.5em]{468\,s}         \\[0.3em]
          \shortstack[l]{LightSB-M\\(2024)}  & \shortstack[c]{0.451\\$\pm$ 0.072}          & \shortstack[c]{0.371\\$\pm$ 0.113}          & \shortstack[c]{0.376\\$\pm$ 0.045}           & \shortstack[c]{0.426\\$\pm$ 0.069}          & \shortstack[c]{0.344\\$\pm$ 0.119}          & \shortstack[c]{0.359\\$\pm$ 0.048}          & \shortstack[c]{\textbf{0.189}\\$\pm$ 0.027} & \shortstack[c]{\textbf{0.176}\\$\pm$ 0.017} & \shortstack[c]{0.244\\$\pm$ 0.070}          & \multirow[c]{1}{*}[0.5em]{2742\,s}        \\
          \bottomrule
        \end{tabularx}
      \end{small}
    }
  \end{subtable}
\end{table}

\section{Experiments}
\label{sec:experiments}
In this section, after summarizing the SB baselines, we benchmark our method in synthetic and real-world settings.

\textbf{Experimental setup.}
We compare QDSB with five bridge baselines: DSB \citep{BortoliTHD21}, DSBM \citep{ShiBCD23}, SF2M \citep{0001MFAZHWB24}, SF2M combined with mini-batch partial OT (mPOT) \citep{NguyenNVPH22}, and LightSB-M \citep{GushchinKBK24}.
DSB is the original diffusion SB method, DSBM replaces the simulation-based training step by a matching objective, SF2M is the simulation-free baseline on which our method builds, SF2M + mPOT strengthens the minibatch coupling step, and LightSB-M is a recent SB solver.
For a fair comparison, all methods use the same backbone models and the same training hyperparameters.
The time measurements are conducted on the same machine (AMD Ryzen 9 7900X 12-Core, NVIDIA GeForce RTX 4090) with only one experiment running at a time.
The Maximum Mean Discrepancy (MMD) values are computed between the groundtruth target samples and the predicted target samples from simulated trajectories of the trained models.
We use the Radial Basis Function Kernel (RBF-Kernel) with its parameter set using the median distance heuristic; the RBF parameter is fixed per dataset.

\begin{figure}[t]
  \makebox[\linewidth][l]{\hspace*{0.15em}\includegraphics{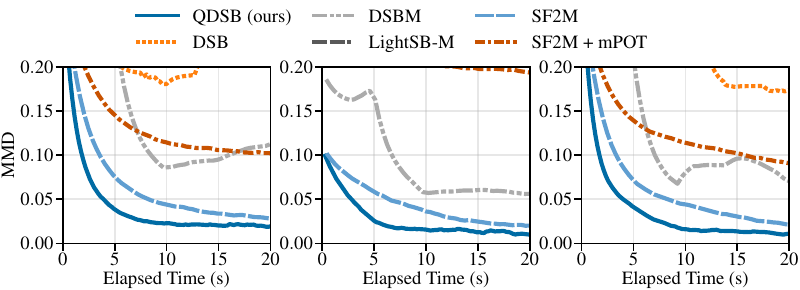}}
  \vspace{0.1em}
  \makebox[\linewidth][l]{\small\hspace{1.1cm} (a) 8Gaussians $\to$ Moons. \hspace{1.33cm} (b) $\,\mathcal{N} \to$ Moons. \hspace{1.7cm} (c) $\,\mathcal{N} \to$ 8Gaussians.}\\[0.5em]
  \makebox[\linewidth][l]{\hspace*{0.15em}\includegraphics{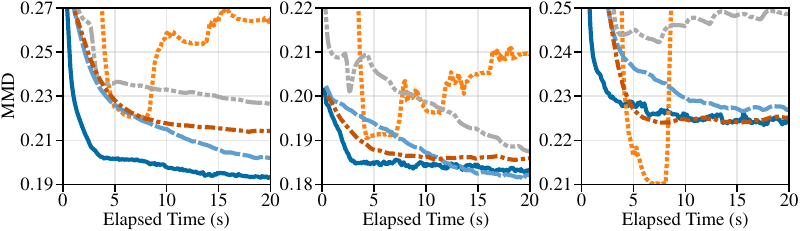}}
  \vspace{0.1em}
  \makebox[\linewidth][l]{\small\hspace{2.25cm} (d) EB. \hspace{3.3cm} (e) Cite. \hspace{3cm} (f) Multi.}
  \caption{Time-quality trade-off curves measured by MMD over wall-clock time.
    Panels~(a)--(c) show the 2D toy datasets, and panels~(d)--(f) show the real-world cellular biology datasets.
    Curves closest to the bottom-left corner indicate the best time-quality trade-off.}
  \label{fig:tradeoff}
\end{figure}

\textbf{2D toy experiments.}
Following related work, we first evaluate all methods on three standard 2D toy settings between the 8Gaussians distribution (blue in Figure~\ref{fig:anchors}), the Moons distribution (green in Figure~\ref{fig:anchors}), and a standard two-dimensional Gaussian distribution.
This yields the three pairwise tasks 8G$\to$M, G$\to$M, and G$\to$8G shown in Table~\ref{tab:results-2d} and Figure~\ref{fig:tradeoff}a--c.
Table~\ref{tab:results-2d} shows that QDSB achieves the lowest MMD after 10\,s and after 60\,s on all three tasks, and it has comparable results after 500 epochs.
Yet, it requires less total wall-clock time compared to most other approaches to complete training.
Figure~\ref{fig:tradeoff}a--c confirms this over the full training horizon: the QDSB curves stay closest to the bottom-left corner in all cases, indicating the best time-quality trade-off among the methods.
We note that DSB fluctuates strongly in some experiments because of alternating between fitting neural networks to the forward and backward process, respectively.

\textbf{Real-world cellular biology experiments.}
We consider three real-world cellular biology datasets from \citet{0001MFAZHWB24}.
EB is an embryoid body differentiation dataset with five population measurements over 30 days, while Cite and Multi are multimodal single-cell benchmarks \citep{LueckenGBCS25}. Following \citet{0001MFAZHWB24}, we use four measurements on days 2, 3, 4, and 7.
Table~\ref{tab:results-rw} shows that QDSB achieves the best MMD after 10\,s on all three datasets, and it remains best on EB and tied best on Cite after 60\,s.
At the longest budget, some baselines obtain slightly lower final MMD values, but only at a substantially higher runtime.
Figure~\ref{fig:tradeoff}d--f shows that QDSB reaches a strong performance faster and therefore provides the best overall time-quality trade-off on these tasks.

\begin{figure}[t]
  \makebox[\linewidth][l]{\hspace*{0.15em}\includegraphics{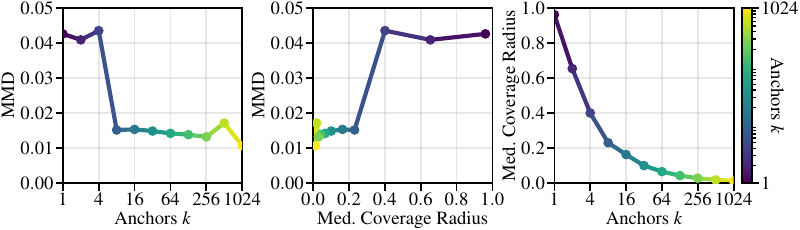}}
  \makebox[\linewidth][l]{\small\hspace{1.1cm} (a) Anchors vs.~MMD. \hspace{1.2cm} (b) Coverage vs.~MMD. \hspace{0.7cm} (c) Anchors vs.\ coverage.}
  \caption{Sensitivity and ablation experiments on the 8Gaussians to Moons dataset. Each point represents a separate QDSB training run over 1000 epochs, averaged over five different seeds.}
  \label{fig:sensitivity}
\end{figure}

\textbf{Sensitivity and ablation experiments.}
Figure~\ref{fig:sensitivity} illustrates the effect of the number of anchors per marginal on the 8Gaussians-to-Moons task, and reports the resulting MMD (Figure~\ref{fig:sensitivity}a), the median coverage radius of the selected anchors (Figure~\ref{fig:sensitivity}b), and their relation (Figure~\ref{fig:sensitivity}c), averaged over five seeds.
As the number of anchors increases, the MMD improves consistently (except for the slight bump at $k = 512$, which might be an artifact of the random initial anchor selection).
This is consistent with our theory, since a finer anchor quantization yields a more accurate coupling approximation.
Beyond about $k = 16$, however, the gains become smaller, which matches the fact that the coverage radius decreases more slowly afterwards.
The case $k = 1$ also serves as an ablation experiment.
This is because with only one anchor per marginal, all samples lie in a single anchor cell, so the induced coupling is equivalent to a random OT plan not using any geometric structure.
Accordingly, $k \in \{1, 2, 4\}$ gives the worst quality as such few anchors do not convey enough geometry.

\begin{figure}[t]
  \centering
  \includegraphics{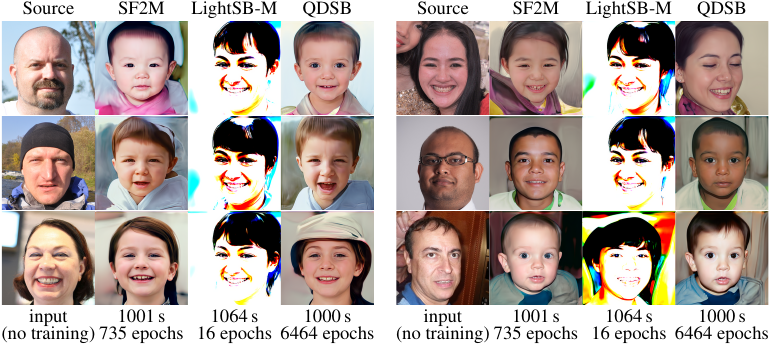}
  \caption{Qualitative comparison of adult-to-child image translation on FFHQ with a fixed time budget of 1000\,s.
    The left column shows adult input images, and the remaining columns show the translated child images.
    Additional examples are provided in Appendix~\ref{sec:additional-experiments}.}
  \label{fig:ffhq_fixed_budget}
\end{figure}

\textbf{Image translation with fixed time budget.}
Following the image-generation experiments of \citet{GushchinKBK24,AlouadiHLMPT26}, we consider image translation on FFHQ \citep{KarrasLA19}, which contains unpaired images of adults and children.
The task is to learn a map between the adult and child distributions.
We encode all images with the ALAE encoder \citep{PidhorskyiAD20} and run all methods in the resulting 512-dimensional latent space.
All methods are trained under the same time budget of 1000\,s.
Figure~\ref{fig:ffhq_fixed_budget} shows the resulting samples.
We were not able to obtain usable results from DSB and DSBM in this setting.
Within the time budget, QDSB completes 6464 epochs, while LightSB-M completed only 16 epochs.
This gap is reflected clearly in the image quality: LightSB-M remains far from a usable translation regime, whereas SF2M and especially QDSB produce coherent images.
Overall, QDSB gives the strongest qualitative results.

\section{Conclusion}
\label{sec:conclusion}
We proposed a quantized coupling framework for simulation-free SB training from unpaired samples.
QDSB replaces repeated minibatch OT solves by a single entropic transport problem on anchor-quantized endpoint distributions and then lifts the resulting anchor‑level transport plan back to the original data via cell-wise sampling.
We provided a theoretical analysis of this approximation, showing that both the entropic transport value and the optimal regularized coupling remain stable under quantization and that anchor coverage directly controls the induced error.
Empirically, experiments on toy benchmarks and cellular biology datasets demonstrate that QDSB preserves the generation quality of existing simulation‑free baselines while substantially improving wall‑clock time needed for training.
Overall, quantized couplings offer a simple, principled, and effective way to alleviate the coupling bottleneck in simulation-free SB training.

\bibliographystyle{plainnat}
\bibliography{references.bib}

\newpage
\appendix

\section{Proofs}
\label{sec:proofs}
This section collects all proofs of our results stated in the main text.

\subsection{Proof of Theorem~\ref{thm:main}}
\paragraph{Part~$(i)$.}
The proof of part~$(i)$ first establishes that $x \mapsto (x, T_i(x))$ describes a coupling. Then, we directly upper-bound the Wasserstein distance.

Let $i \in \{0, 1\}$ and $\Phi_i : \R^d \to \R^d \times \R^d$ with $\Phi_i(x) := (x, T_i(x))$.
Since $T_i$ is measurable, $\Phi_i$ is measurable.
Define the measure $\gamma_i := (\Phi_i)_{\#} q_i$ on $\R^d \times \R^d$ equipped with its Borel sigma-algebra, and let $\zeta_x(x, y) := x$ and $\zeta_y(x, y) := y$ denote the coordinate maps.
Then,
\begin{align}
  (\zeta_x)_{\#} \gamma_i & = (\zeta_x)_{\#} (\Phi_i)_{\#} q_i = (\zeta_x \circ \Phi_i)_{\#} q_i = (\mathrm{Id}_{\R^d})_{\#} q_i = q_i \text{, and} \\
  (\zeta_y)_{\#} \gamma_i & = (\zeta_y)_{\#} (\Phi_i)_{\#} q_i = (\zeta_y \circ \Phi_i)_{\#} q_i = (T_i)_{\#} q_i = \tilde{q}_i \text{.}
\end{align}
Therefore, $\gamma_i$ is a coupling between $q_i$ and $\tilde{q}_i$, that is, $\gamma_i \in \Pi(q_i, \tilde{q}_i)$.
Then, we discern two cases.
\begin{itemize}
  \item For $a < \infty$, we have
        \begin{align}
          \MoveEqLeft W_a(q_i, \tilde{q}_i)^a
          \overset{(*)}{=} \inf_{\pi \in \Pi(q_i, \tilde{q}_i)} \int \left\| x - y \right\|^a \mathrm{d} \pi(x, y)
          \overset{(**)}{\leq} \int \left\| x - y \right\|^a \mathrm{d} \gamma_i(x, y)                                                        \\
           & \overset{(\mathord{*}\mathord{*}\mathord{*})}{=} \int \left\| x - T_i(x) \right\|^a \mathrm{d} q_i(x) = \varepsilon_i^a \text{,}
        \end{align}
        where $(*)$ uses the definition of the Wasserstein metric, $(**)$ upper-bounds the infimum with the coupling $\gamma_i$, and $(\mathord{*}\mathord{*}\mathord{*})$ uses the definition of $\Phi_i$ and the change-of-variables formula.
        Since all terms are non-negative, taking the $a$-th root gives $W_a(q_i, \tilde{q}_i) \leq \varepsilon_i$.

  \item Similarly, for $a = \infty$, we have
        \begin{align}
          W_\infty(q_i, \tilde{q}_i)
          = \inf_{\pi \in \Pi(q_i, \tilde{q}_i)} \esssup_{(x, y) \sim \pi} \left\| x - y \right\|
          \leq \esssup_{(x, y) \sim \gamma_i} \left\| x - y \right\|
          = \esssup_{x \sim q_i} \left\| x - T_i(x) \right\| = \varepsilon_i \text{.}
        \end{align}
\end{itemize}

\paragraph{Part~$(ii)$.}
The proof proceeds by reducing $\obj_\tau$ to the normalization used by \citet{EcksteinN22} and then applying their value-stability theorem (recalled in Theorem~\ref{thm:eck-3-7}).

Define the rescaled cost function $\bar{c} := \frac{1}{\tau} c$ and the rescaled objective function
\begin{align}
  \bar{\obj}(\alpha, \beta) := \inf_{\pi \in \Pi(\alpha, \beta)} \int \bar{c} \,\mathrm{d} \pi(x, y) + \KL(\pi \| \alpha \otimes \beta) \text{,}
\end{align}
for any $\alpha, \beta \in \probs_a(\R^d)$.
Since $\tau > 0$, we have $\obj_\tau(\alpha, \beta) = \tau \bar{\obj}(\alpha, \beta)$ because the positive constant $\tau$ commutes with taking the infimum.

We next show that the assumed admissibility of $c$ with constant $L$ implies that of $\bar c$ with constant $L / \tau$.
Indeed, let $r_0, r'_0, r_1, r'_1 \in \probs_a(\R^d)$, $\pi \in \Pi(r_0, r_1)$, and $\pi' \in \Pi(r'_0, r'_1)$. Then,
\begin{align}
  \left| \int \bar{c} \,\mathrm{d}(\pi - \pi') \right|
  \overset{(*)}{=} \left| \int \frac{1}{\tau} c \,\mathrm{d}(\pi - \pi') \right|
  \overset{(**)}{=} \frac{1}{\tau} \left| \int c \,\mathrm{d}(\pi - \pi') \right|
  \overset{(\mathord{*}\mathord{*}\mathord{*})}{\leq} \frac{L}{\tau} W'_a(\pi, \pi') \text{,} \label{eq:bar-c-admissibility-constant}
\end{align}
where $(*)$ holds by the definition of $\bar{c}$, $(**)$ by the homogeneity of the integral, and $(\mathord{*}\mathord{*}\mathord{*})$ uses Assumption~\ref{thm:a1}.

Define the marginal perturbation size as
\begin{align}
  \delta_a := \begin{cases}
                \left( W_a(q_0, \tilde{q}_0)^a + W_a(q_1, \tilde{q}_1)^a \right)^{1/a} & \text{for } a < \infty \text{,} \\
                \max \{ W_\infty(q_0, \tilde{q}_0), W_\infty(q_1, \tilde{q}_1) \}      & \text{for } a = \infty \text{.}
              \end{cases}
\end{align}
By part~$(i)$, $W_a(q_0, \tilde{q}_0) \leq \varepsilon_0$ and $W_a(q_1, \tilde{q}_1) \leq \varepsilon_1$.
Therefore, if $a < \infty$,
\begin{align}
  \delta_a = \left( W_a(q_0, \tilde{q}_0)^a + W_a(q_1, \tilde{q}_1)^a \right)^{1/a} \overset{(*)}{\leq} (\varepsilon_0^a + \varepsilon_1^a)^{1/a} = \Delta_a \text{,}
\end{align}
where $(*)$ holds as $x \mapsto x^a$ and $x \mapsto x^{1/a}$ are non-decreasing on $[0, \infty)$.
If $a = \infty$,
\begin{align}
  \delta_\infty = \max \{ W_\infty(q_0, \tilde{q}_0), W_\infty(q_1, \tilde{q}_1) \} \overset{(**)}{\leq} \max \{ \varepsilon_0, \varepsilon_1 \} = \Delta_\infty \text{,}
\end{align}
where $(**)$ holds by the properties of the maximum.

We now apply Theorem~\ref{thm:eck-3-7} with marginals $(q_0, q_1)$ and $(\tilde{q}_0, \tilde{q}_1)$ and cost $\bar{c}$.
Since $\bar{c}$ satisfies the admissibility condition with constant $L / \tau$, the theorem yields
\begin{align}
  \left| \bar{\obj}(q_0, q_1) - \bar{\obj}(\tilde{q}_0, \tilde{q}_1) \right| \leq \frac{L}{\tau} \delta_a \text{.}
\end{align}

Putting all steps together, we obtain
\begin{align}
  \left| \obj_\tau(q_0, q_1) - \obj_\tau(\tilde{q}_0, \tilde{q}_1) \right| = \tau \left| \bar{\obj}(q_0, q_1) - \bar{\obj}(\tilde{q}_0, \tilde{q}_1) \right| \leq \tau \frac{L}{\tau} \delta_a \leq L \Delta_a \text{.}
\end{align}

\paragraph{Part~$(iii)$.}
Recall $\bar{c}$, $\bar{\obj}$, and $\delta_a$ from the proof of part~$(ii)$.
Since scaling the objective by a positive constant does not change its minimizers, $\pi^\tau_{q_0,q_1}$ and $\pi^\tau_{\tilde{q}_0,\tilde{q}_1}$---assumed optimal for $\obj_\tau$---are the optimal couplings of $\bar{\obj}$.

We now apply Theorem~\ref{thm:eck-3-11} with marginals $(q_0, q_1)$ and $(\tilde{q}_0, \tilde{q}_1)$, cost $\bar{c}$, admissibility constant $\bar{L} := L / \tau$ (established in \eqref{eq:bar-c-admissibility-constant}), and the transport inequality constant $C_b$ from Assumption~\ref{thm:a2}.
This yields
\begin{align}
  \MoveEqLeft W'_{b}(\pi^\tau_{q_0,q_1}, \pi^\tau_{\tilde{q}_0,\tilde{q}_1})
  \overset{(*)}{\leq} 2^{1/b - 1/a} \delta_a + C_{b}(2 \bar{L} \delta_a)^{1/(2b)}
  \overset{(**)}{=} 2^{1/b - 1/a} \delta_a + C_{b}(2 L \delta_a / \tau)^{1/(2b)}                                               \\
   & \overset{(\mathord{*}\mathord{*}\mathord{*})}{\leq} 2^{1/b - 1/a} \Delta_a + C_{b}(2 L \Delta_a / \tau)^{1/(2b)} \text{,}
\end{align}
where $(*)$ applies Theorem~\ref{thm:eck-3-11}, $(**)$ substitutes $\bar{L}$, and $(\mathord{*}\mathord{*}\mathord{*})$ holds by the intermediate results in the proof of part (ii) and as the map
\begin{align}
  x \mapsto 2^{1/b - 1/a} x + C_{b}(2 L x / \tau)^{1/(2b)}
\end{align}
is non-decreasing on $[0, \infty)$.
This completes the proof of part~$(iii)$.

\subsection{Proof of Lemma~\ref{thm:radius}}
\paragraph{Part~$(i)$.}
By assumption, for each $i \in \{0, 1\}$, $\|x_i - T_i(x_i)\| \leq r_i$ for $q_i$-almost every $x_i$.
Then, we discern two cases.
\begin{itemize}
  \item For $a < \infty$, we have
        \begin{align}
          \varepsilon_i^a
          \overset{(*)}{=} \int \|x_i - T_i(x_i)\|^a \, \mathrm{d} q_i(x_i)
          \overset{(**)}{\leq} \int r_i^a \, \mathrm{d} q_i(x_i)
          \overset{(\mathord{*}\mathord{*}\mathord{*})}{=} r_i^a \text{,}
        \end{align}
        where $(*)$ substitutes the definition of $\varepsilon_i$, $(**)$ uses the assumption and that $x \mapsto x^a$ is non-decreasing on $[0, \infty)$, and $(\mathord{*}\mathord{*}\mathord{*})$ simplifies the constant in the integral.

  \item For $a = \infty$, we have
        \begin{align}
          \varepsilon_i
          \overset{(*)}{=} \esssup_{x_i \sim q_i} \|x_i - T_i(x_i)\|
          \overset{(**)}{\leq} r_i \text{,}
        \end{align}
        where $(*)$ substitutes the definition of $\varepsilon_i$ and $(**)$ directly uses the assumption.
\end{itemize}

\paragraph{Part~$(ii)$.}
Similar to the proof of part~$(i)$, we discern two cases.
\begin{itemize}
  \item For $a < \infty$, we have
        \begin{align}
          \Delta_a^a
          \overset{(*)}{=} \varepsilon_0^a + \varepsilon_1^a
          \overset{(**)}{\leq} r_0^a + r_1^a \text{,}
        \end{align}
        where $(*)$ substitutes the definition of $\Delta_a$ and $(**)$ uses part~$(i)$ and that $x \mapsto x^a$ is non-decreasing on $[0, \infty)$.

  \item For $a = \infty$, we have
        \begin{align}
          \Delta_\infty
          \overset{(*)}{=} \max \{ \varepsilon_0, \varepsilon_1 \}
          \overset{(**)}{\leq} \max \{ r_0, r_1 \} \text{,}
        \end{align}
        where $(*)$ substitutes the definition of $\Delta_a$ and $(**)$ follows by the properties of the maximum.
\end{itemize}

\section{External theorems} \label{sec:external-theorems}
This section collects theorems that we use in the proofs of our statements.

\begin{theorem}[\citealt{EcksteinN22}, Theorem~3.7~$(ii)$, $N = 2$]
  \label{thm:eck-3-7}
  Let $a \in [1, \infty]$, $r_0, r'_0, r_1, r'_1 \in \probs_a(\R^d)$, the cost function $c$ admissible with constant $L \geq 0$ (Assumption~\ref{thm:a1}), and
  \begin{align}
    \bar{\obj}(\alpha, \beta) := \inf_{\pi \in \Pi(\alpha, \beta)} \left[ \int c \,\mathrm{d} \pi(x, y) + \KL(\pi \| \alpha \otimes \beta) \right]
    \text{.} \label{eq:eckstein-obj}
  \end{align}
  Then,
  \begin{align}
    | \bar{\obj}(r_0, r_1) - \bar{\obj}(r'_0, r'_1) | \leq L \delta_a \text{, where }
    \delta_a := \begin{cases}
                  \left( W_a(r_0, r'_0)^a + W_a(r_1, r'_1)^a \right)^{1/a} & \text{for } a < \infty \text{,} \\
                  \max \{ W_\infty(r_0, r'_0), W_\infty(r_1, r'_1) \}      & \text{for } a = \infty \text{.}
                \end{cases}
    \quad\label{eq:eck-delta}
  \end{align}
\end{theorem}

\begin{theorem}[\citealt{EcksteinN22}, Theorem~3.11, $N = 2$]
  \label{thm:eck-3-11}
  Let $a \in [1, \infty]$, $b \in [1, \infty)$ with $b \leq a$, $r_0, r'_0, r_1, r'_1 \in \probs_a(\R^d)$. Suppose that $(r_0, r_1)$ satisfy Assumption~\ref{thm:a2} with constant $C_b$, and that the cost function $c$ satisfies Assumption~\ref{thm:a1}.
  Recall $\bar{\obj}$ from~\eqref{eq:eckstein-obj} and $\delta_a$ from~\eqref{eq:eck-delta}.
  Then, the optimizers $\pi_{r_0,r_1}, \pi_{r'_0,r'_1}$ of $\bar{\obj}(r_0,r_1)$ and $\bar{\obj}(r'_0,r'_1)$, respectively, satisfy
  \begin{align}
     & W'_b(\pi_{r_0,r_1}, \pi_{r'_0,r'_1}) \leq 2^{1/b-1/a} \delta_a + C_b (2 L \delta_a)^{1/(2b)} \text{.}
  \end{align}
\end{theorem}

The next result provides guarantees on the farthest-first traversal approach to the $k$-center problem, recalled in Algorithm~\ref{alg:farthest-first} for self-completeness.

\begin{algorithm}[ht]
  \caption{Greedy $k$-center: Farthest-first traversal.}
  \label{alg:farthest-first}
  \begin{algorithmic}[1]
    \item[\textbf{Input:}] Finite point set $\{x^{(i)}\}_{i=1}^n \subseteq \R^d$, anchor set cardinality $k \in \{1, \dots, n\}$;
    \item[\textbf{Output:}] Anchor set $A \subseteq \{x^{(i)}\}_{i=1}^n$ with $|A| = k$;
    \vspace{0.2em}
    \STATE Select an arbitrary initial point $x_{\mathrm{init}} \in \{x^{(i)}\}_{i=1}^n$ and set $A \leftarrow \{x_{\mathrm{init}}\}$;
    \WHILE{$|A| < k$}
    \STATE $x_{\mathrm{next}} \leftarrow \arg\max_{x \in \{x^{(i)}\}_{i=1}^n} \min_{\alpha \in A} \|x - \alpha\|$;
    \STATE $A \leftarrow A \cup \{x_{\mathrm{next}}\}$;
    \ENDWHILE
    \STATE \textbf{return} $A$;
  \end{algorithmic}
\end{algorithm}

\begin{theorem}[\citealt{kmetric}, Theorem~4.3]
  \label{thm:kmetric}
  Given a finite set of points $V \subseteq \R^d$ and number of anchors $k$, the goal of the $k$-center clustering problem is to find a set $C \subseteq V$ with $|C| = k$ that minimizes the associated coverage radius
  \begin{align}
    r_C = \max_{u \in V} \min_{v \in C} c(u, v) \text{.}
  \end{align}
  The optimal coverage radius is denoted by $r^*$.
  The greedy $k$-center algorithm in Algorithm~\ref{alg:farthest-first} is a 2-approximation to the optimal $k$-center clustering problem,
  that is, $r_A \leq 2 r^*$, where $r_A$ uses the anchor set $A$ output by the algorithm.
\end{theorem}

\section{Additional experimental results}
\label{sec:additional-experiments}
\begin{figure}[tp]
  \centering
  \includegraphics{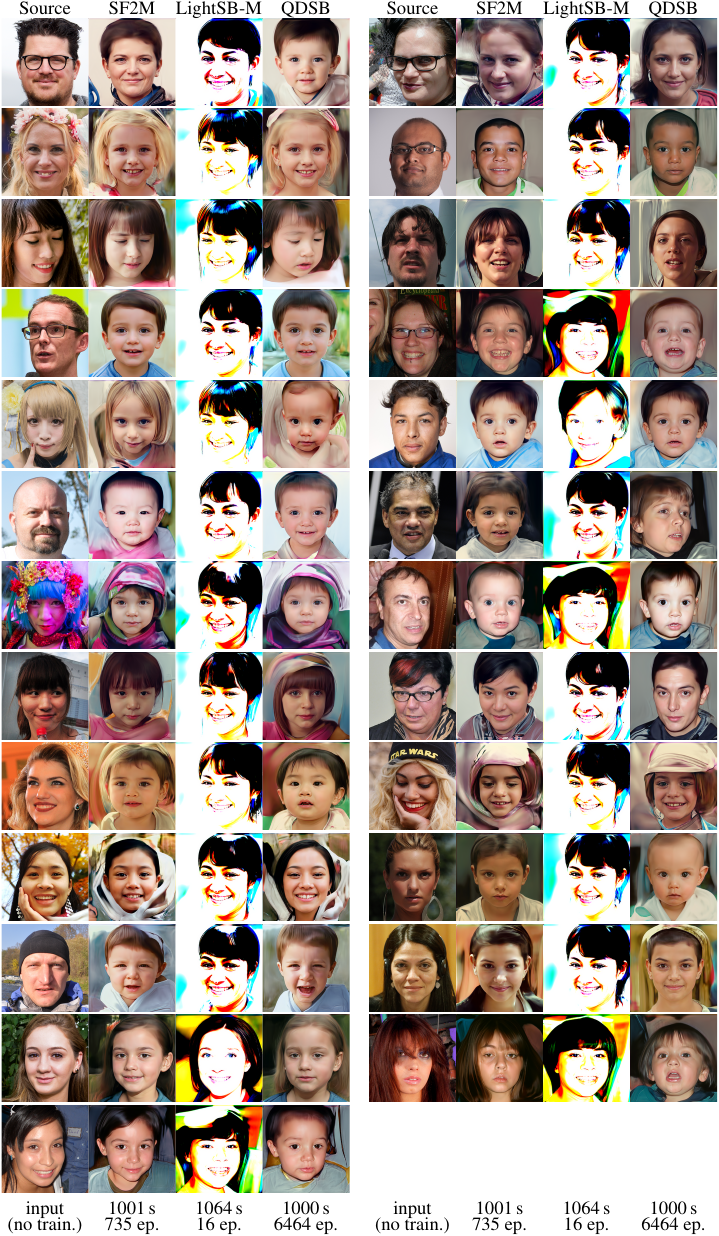}
  \caption{Additional qualitative results for the fixed-budget FFHQ experiment from Figure~\ref{fig:ffhq_fixed_budget}.
    Each row shows an adult input together with the corresponding outputs of SF2M, LightSB-M, and QDSB under the same training-time budget.}
  \label{fig:imgfull1}
\end{figure}

\begin{figure}[tp]
  \centering
  \includegraphics{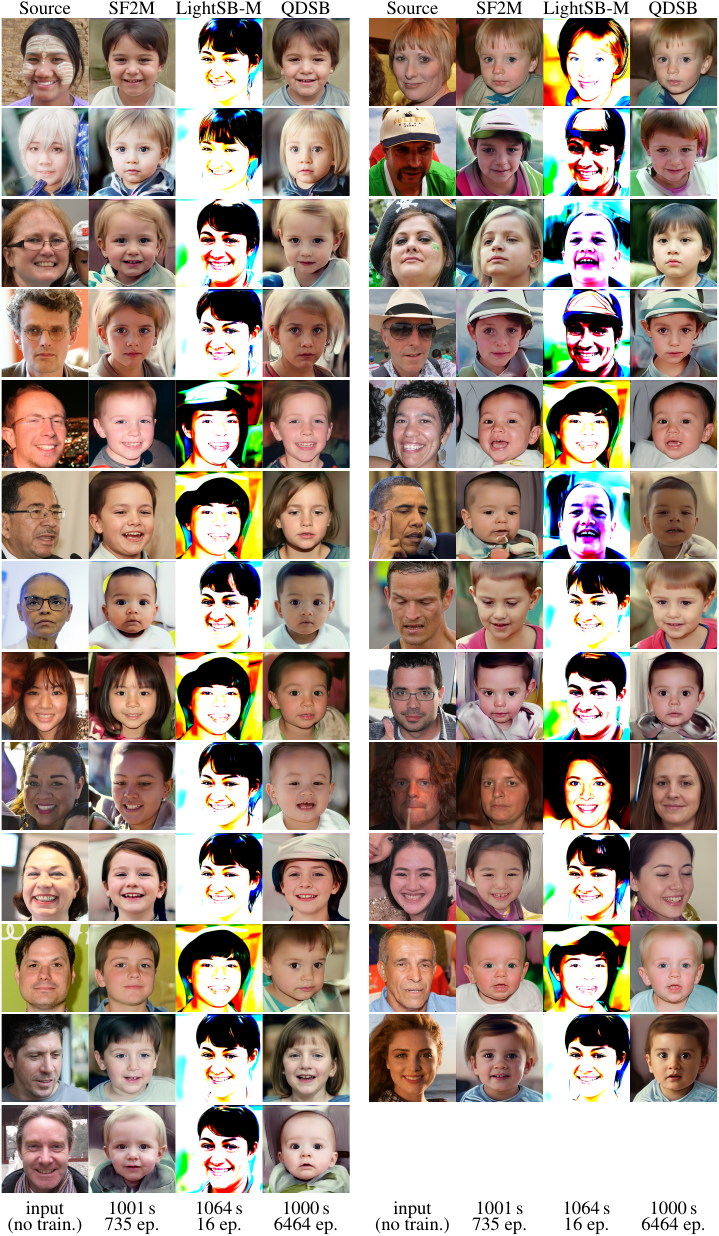}
  \caption{More qualitative results for the fixed-budget FFHQ experiment from Figure~\ref{fig:ffhq_fixed_budget}.
    The same pattern as in Figure~\ref{fig:imgfull1} can be observed across additional examples.}
  \label{fig:imgfull2}
\end{figure}

Figures~\ref{fig:imgfull1} and~\ref{fig:imgfull2} provide additional qualitative examples for the fixed-budget FFHQ experiment from Figure~\ref{fig:ffhq_fixed_budget}.
They use the same adult-to-child translation task, the same 512-dimensional ALAE latent space, and the same wall-clock budget of 1000\,s.
As in the main figure, we only report SF2M, LightSB-M, and QDSB because we were not able to obtain usable results for DSB and DSBM.
The qualitative trend is unchanged: QDSB produces the most consistent child-face translations, SF2M often gives reasonable outputs, and LightSB-M remains unstable given this budget.


\section{Additional experimental details}
\label{sec:experimental-details}
This section augments Section~\ref{sec:experiments} by summarizing the main training details and hyperparameter choices of all reported methods.

\textbf{Common setup.}
All code and data are available at \gh.
We ran the experiments on a machine with an AMD Ryzen 9 7900X CPU and an NVIDIA GeForce RTX 4090 GPU.
The toy and single-cell benchmarks were run on CPU, whereas the FFHQ latent-space experiment was run on GPU.
For the toy, single-cell, and sensitivity results we report averages over five seeds $\{0,1,2,3,4\}$, while the FFHQ image experiment is qualitative and uses a single seed $\{0\}$.
Whenever time-quality curves are recorded, elapsed time is measured with evaluation overhead removed, so Figures~\ref{fig:tradeoff} and~\ref{fig:sensitivity} reflect training time rather than checkpoint time.
In the toy suite, the curves are averaged over seeds, and in the single-cell benchmarks they are averaged over seeds and leave-one-out tasks.
For the quantitative experiments, quality is measured by MMD with an RBF kernel whose bandwidth is chosen by the median heuristic on 4096 reference points.
For the toy and single-cell benchmarks, final MMD is computed on the full evaluation populations.
All experiments were run sequentially to avoid interference.
Running all experiments takes about one week on the hardware described above.

\textbf{Model architecture and optimization.}
Across all methods, we use the same time-dependent four-layer MLP backbone.
The layers are $d-64-64-d$, where $d$ is the input dimension and the hidden layers use a width of 64.
We use the AdamW optimizer with learning rate $10^{-4}$ and weight decay $10^{-2}$, batch size 256, and rollout batch size 2048.
LightSB-M uses its standard Gaussian-mixture parameterization.
In the FFHQ latent experiment, all methods use the AdamW optimizer with learning rate $10^{-3}$ and weight decay $10^{-2}$, batch size 512, and rollout batch size 2048.

\textbf{2D toy experiments.}
The three tasks are 8Gaussians$\to$Moons, Gaussian$\to$Moons, and Gaussian$\to$8Gaussians.
Each endpoint distribution provides 16384 training samples and 4096 evaluation samples.
In the main toy benchmark, all methods are trained for 500 epochs and MMD is recorded every epoch.
We use an exact numerical OT solver with squared Euclidean cost, $\sigma=0.25$, and 100 Euler-Maruyama steps per unit time.
Our approach uses 256 anchors per marginal, farthest-first anchor selection, anchor refresh every 100 epochs, and random sampling within matched anchor cells.
SF2M+mPOT differs from SF2M only in the endpoint coupling and uses a partial transport fraction of 0.8.
LightSB-M uses 128 Gaussian components and a minimum covariance scale of $10^{-2}$.
For DSB and DSBM, we choose the inner iteration count automatically so that the total number of parameter updates matches the SF2M update budget.
In the reported toy runs, DSB uses 20 IPF rounds, 20 Langevin steps, and constant step size $\gamma = 0.01$.
DSBM uses 20 outer iterations, $\sigma=0.25$, 100 rollout steps per unit time, and loss weighting.
The sensitivity experiment in Figure~\ref{fig:sensitivity} keeps the QDSB setup fixed, varies the number of anchors from 1 to 1024, and trains for 1000 epochs for each of the anchor counts.

\textbf{Single-cell experiments.}
For EB, we use the five observed timepoints with nominal times $0,1,2,3,4$ and leave out the three intermediate timepoints.
For Cite and Multi, we use donor 13176 at days $2,3,4,7$ and leave out days 3 and 4.
All methods operate on a 5-dimensional representation obtained by fitting an incremental PCA with 100 components on the raw data, whitening the resulting representation, and then truncating to five dimensions, following the setup of \citet{0001HWDK20}.
All single-cell methods are trained for 1000 epochs and MMD is recorded every epoch.
We use an exact numerical OT solver with squared Euclidean cost, discrete model times, and 100 Euler-Maruyama steps per unit time.
The bridge noise values are 0.25 for EB and Cite, and 1.0 for Multi.
Our approach uses 256 anchors per marginal, farthest-first anchor selection, anchor refresh every 100 epochs, and random endpoint sampling within cells.
SF2M+mPOT again differs from SF2M only in the coupling sampler and uses a partial transport fraction of 0.8.
LightSB-M uses 512 Gaussian components and a minimum covariance scale of $10^{-2}$.
For DSB and DSBM, we again choose the inner iteration count automatically so that the total number of parameter updates matches the SF2M update budget.
In the reported single-cell runs, DSB uses 20 IPF rounds, 20 Langevin steps, and a constant step size of $\gamma = 0.01$.
DSBM uses 20 outer iterations, 20 rollout steps per unit time, and loss weighting.

\textbf{Image translation experiment.}
We use the FFHQ data split into adult age groups 20--29, 30--39, and 40--49 and child age groups 0--2, 3--6, and 7--9, with evaluation fraction 0.1 and split seed 0.
The images are encoded using a pretrained ALAE checkpoint (\texttt{model\_157.pth}), which gives a 512-dimensional latent representation.
All reported image runs use unpaired adult and child latents, a fixed wall-clock budget of 1000\,s, and 50 decoded examples per method.
All methods use the same 512-dimensional MLP backbone described above.
Our approach uses 512 anchors per marginal, no anchor refresh, and random endpoint sampling within cells.
LightSB-M uses 10 Gaussian components and a minimum covariance scale of $10^{-2}$.
We also ran DSB and DSBM in this setting, but they did not produce usable image translations and are therefore omitted from the qualitative figures.
For completeness, the image-space DSB configuration uses 50 inner iterations, 20 IPF rounds, 20 Langevin steps, and a constant step size of $\gamma = 0.01$.
The image-space DSBM configuration uses hidden dimension 512, 50 inner iterations, 20 outer iterations, 100 rollout steps per unit time, and loss weighting.

\end{document}